# Prediction of Industrial Process Parameters using Artificial Intelligence Algorithms

Abdelmoula Khdoudi
*Artificial Intelligence for Engineering Sciences Team - IASI*
*ENSAM-University My ISMAIL*
*Meknes, Morocco*
khdoudi.ma@gmail.com   Abdelmoula.khdoudi@eu.agc.com

Tawfik Masrour
*Artificial Intelligence for Engineering Sciences Team - IASI*
*ENSAM-University My ISMAIL*
*Meknes, Morocco*
t.masrour@ensam.umi.ac.ma

*Abstract*—In the present paper, a method of defining the industrial process parameters for a new product using machine learning algorithms will be presented. The study will describe how to go from the product characteristics till the prediction of the suitable machine parameters to produce a good quality of this product, and this is based on an historical training dataset of similar products with their respective process parameters. In the first part of our study, we will focus on the ultrasonic welding process definition, welding parameters and on how it operate. While in second part, we present the design and implementation of the prediction models such multiple linear regression, support vector regression, and we compare them to an artificial neural networks algorithm. In the following part, we present a new application of Convolutional Neural Networks (CNN) to the industrial process parameters prediction. In addition, we will propose the generalization approach of our CNN to any prediction problem of industrial process parameters. Finally the results of the four methods will be interpreted and discussed.

## I. INTRODUCTION

### A. Motivation

The actual study present the application of machine learning algorithms in the manufacturing field in order to save time during the setting of machines and process parameters for a new type of product. Generally, this activity is conducted by process engineers and it is based on previous experience, product similarity study and comparison with old products, physical tests of several values and the correlation of parameters with the quality output of the desired product.

In this paper we will develop and compare, based on the product characteristics as input data and process parameters values as output, several machine learning models that will help on predicting the best parameters value for a new product. The model will learn how the parameters of the machine changes according to different type of products (historical dataset) and then, will be able to estimate a new parameters value for a new set of characteristics for a new coming product.

### B. Related work

In *Prediction of Best Combination of Process Parameters for Petonation Gun Coating Process Through Taguchi Technique,* K.N.Balan, et al., experimented the optimization of D-spray coating process parameters using Taguchi method, and this in order to find the best processing conditions and to get higher quality of coating. They managed to define very few experiments depended on the number and level of each factor.

In *Prediction of Optimal Process Parameters for Abrasive Assisted Drilling of SS304,* Kapil Kumar, et al., presented a cutting parameters optimization study based on factorial design of response methodology (RSM) in order to improve the surface finish of stainless steel SS304 in the abrasive assisted drilling. They carried out an analysis of variance in order to find out the significance and percentage contribution of process parameters. They reached an overall improvement of 10.81% in surface finish by optimizing the spindle speed, feed rate, and slurry concentration.

Also, RSM methodology was used by B.Vijaya Sankar, et al., in their work *Prediction of Spot Welding Parameters for Dissimilar Weld Joints.* They presented a study on how to reach a desired mechanical properties of spot weld which are the Tensile Strength and the Hardness by optimizing the Electrode force, Weld Current and the Weld time.

Finally, in *Intelligent Prediction of Process Parameters for Bending Forming,* Shengle Ren, et al., introduced a machine learning technique to the concept of process parameters prediction. They experimented mainly the Artificial Neural Networks for the prediction of the pipe forming process parameters which are the bending moment and the boost power. They considered twelve ANN inputs which are mainly related to the pipe characteristics, and they reached an error value that is under 2%.

In our present study, we experiment four machine learning algorithms in order to compare the results of each model and its accuracy for each parameter. Also, we will introduce a new approach of predicting the process parameters which is the using of Convolutional Neural Networks. The value add of using ML models is to avoid any physical experiments and save material and time. This advantage is not present in the classical optimization methods.

## II. THE ULTRASONIC WELDING PROCESS

### A. The Ultrasonic Welding System

The generator (Power Supply): it sends an alternating current whose frequency corresponds to the vibration sought of the welding. The converter (or transducer) which is composed from piezoelectric ceramics: it transforms the alternating current into mechanical vibrations [3]. The booster: Due to their mechanical resonance frequency, they allow to mechanically vary the amplitude of the vibration. The sonotrode: it is the ultimate element of the chain (Fig.1) that transmits the produced vibration and thus allows the transfer of energy.

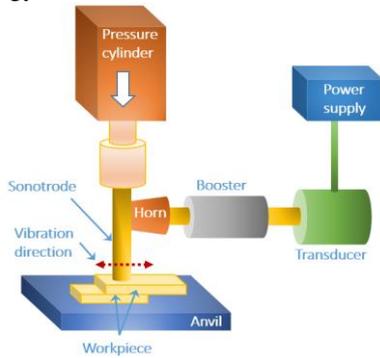

Fig. 1.  The Ultrasonic Welding System

### B. The Ultrasonic Welding Parameters

*1) The Energy and the Welding Time*

The welding energy propagates through the material (copper) for a certain time to ensure the weldability of the node. Depending on the vibration amplitude of the sonotrode, the welding pressure and the quality of the wires, the welding time varies between 0.2 and 1.5 seconds. During the welding operation and through the first contact phase between the wires and the welding parts - compression process - a time of at least 0.2 seconds is required. A slow welding time (more than 1.5 seconds) can cause overheating, damage to the ultrasonic nodes and a significant reduction in the service life of the wear parts.

*2) Welding Pressure*

During the welding operation, the sonotrode apply high frequency vibrations to the workpiece in parallel to a working pressure that is driven by pneumatic force. The pressure ensure a good mechanical adhesion and welding point compression [3]. The welded point strength increase proportionally by the pressure increase but passing a certain limit, some defects can be observed such as node burn, welding burr or even material structure damage.

*3) Welding Amplitude*

The amplitude represent the upward-downward displacement of horn during the application of high frequency vibration. The square value of the amplitude gives the heating quantity generated at the contact surface sonotrode/piece. With higher amplitude value, the higher is the impact of friction and then the better is the weldability.

### C. The Welding of Electrical Copper wires

In this application, the output product of the welding process is a set of wires welded together (different sections), the welding points must be consistent : resistant to a certain breaking force defined by the customer, not burned and without burrs [5]. Due to dimensional limitations of the used machine, the number of wires that can be welded is fixed to maximum five wires per side for points in bilateral welding or fifteen wires on one side (unilateral node) (Fig.2).

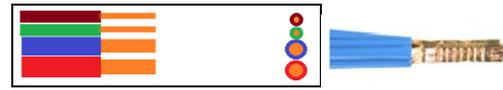

Fig. 2.  Layout of a welding node (unilateral or bilateral)

The wires produced within the factory are in different sections. The number of combinations that we can compose by playing on the number of wires per node and each wire section is huge.

### D. Setting the Process Parameters for a New Product.

The actual approach used in real production to set the ultrasonic welding parameters follows some heuristic steps. The user define some random values of Energy, Pressure and Amplitude (generally based on personal experience), then perform the welding operation for the new product. The node is inspected visually and tested on the pull force machine to check its pull force resistance. Results are rarely positive since the first trial, that means the machine user adjust the parameters value several times to reach the requested pull force resistance and the aspect conformance.

The procedure of searching the best parameters is actually costly because some major loses are unavoidable such as machine's energy consumed during the tests, material rejects after pull force test for each sample, time lose for one or two persons spending at least 20 minutes per new product (in average 2.5 tests are performed for each new part), a new customer project can contains 80 new product.

## III. DATA PREPARATION AND WORK METHOLOGY

The goal of the study is to develop a model which predict with acceptable accuracy 'and without the need of physical tests' the values of Pressure, Energy and Amplitude that leads to a quality output product, and this, based only on raw-material characteristics. The best candidate to develop such efficient model is to use Machine Learning algorithms, that was demonstrated to predict with high accuracy, new outputs and decisions by learning the hidden features in existing data [6][7][8][9][10]. In our study, the Supervised Learning Methods will be deployed to explore an existing process/product dataset.

### A. Data Preparation

The dataset was issued from the Ultrasonic welding service of an electrical harness production factory. It contains brut data on existing products (currently in production). In some form, the product characteristics was reported in the dataset in addition to their corresponding ultrasonic parameters.

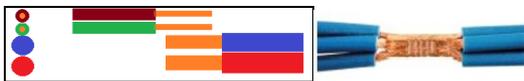

Fig. 3.  Data structure as received from the company

Fig. 4. Data structure after manipulation and cleaning

## B. Conduct of the Study

For any new product, a new set of parameters should be defined in order to get a good welding result. Since we are making the welding operation on the same machine with the same operator and the same row material type, then the parameters are a function of the product design (characteristics). Based on our product and process knowledge, it was not complicated to make a first analysis to select some first set of characteristics suspected to lead to parameters changes. This decision was also confirmed by checking the dataset values of Energy, Pressure and Amplitude.

By this, the selected product characteristics, that will be considered as prediction model input are the cross section value for each wire in both welding point sides.

The maximum number of wires that can be welding is five per side for bilateral point and fifteen in one side for unilateral point. According to the physical tests, for the same number and section of wires, the parameters are the same. For this fact, our input table will consider only the number of wires without taking in consideration the side. In the next two sections, we will perform predictions based on five models which are multiple linear regression, support vector regression, artificial neural networks and finally an introduction of convolutional neural networks application to the field of multi-output regression. The 3D output parameters are in different scales, in addition, we cannot predict them separately because they are dependent, which mean, we should know which combination of Energy, Amplitude and Pressures values are suitable for a new product. This cannot be guaranteed if we predict each value independently. However, to compare the accuracy of different models, we will evaluate the loss value of each parameter separately. Since it is not possible to calculate the accuracy of a linear continuous output, and in order to simulate the prediction accuracy of the used models, we will refer to real process limits. The lower and upper tolerance of ±15% of each welding parameters included in the test data should not be exceeded. That means, if the predicted value is included in the range of tolerance (±15% of the real value), we considered it as correct prediction, and as wrong prediction if the value is out of this process tolerance. This limit is fixed based on company experience, that's mean if a change of less than 15% of any process parameter, the quality result is not negatively impacted.

## IV. PREDICTION OF PROCESS PARAMETERS USING MACHINE LEARNING ALGORITHMS

In this section, we will present the predictive models that we used for our study. The selected methods (Multiple Linear Regression and Support Vector Regression) showed better results on our case study during our pre-tests. Then we decided to compare them with the Artificial Neural Network algorithm and to select the one which give a better prediction results on the validation data.

### A. Multiple Linear Regression Model

The multiple regression models are mathematical models used in many situations to study the association between input data (exploratory factors) and a variable to explain, this can be privileged for a description purpose and / or for a prediction purpose as it's the case for our study [11] [12].

The output variables of our model are the process parameters Energy, Amplitude and Pressure, which are dependent variables and should be predicted in parallel, this is a characterization of Multi-Outputs Regression.

Using existing libraries of multi-output variables regression, we managed to predict simultaneously the three parameters of the ultrasonic welding process for a completely new input values which are the cross section of each wire in the welding node, and based on the prediction table (Tab.I), we will calculate the accuracy and loss for each parameters separately to have a better overview about the model behavior in respect to each variable, Below is a part of the prediction result for this model (Fig.5), following the process tolerance limit.

TABLE I. PREDICTED PARAMETERS USING MULTIPLE LINEAR REGRESSION MODEL

| Energy (Ws) | Amplitude (%) | Pressure (Bar) |
|---|---|---|
| 241.16 | 68.08 | 1.75 |
| 663.17 | 83.18 | 2.49 |
| 212.32 | 67.52 | 1.69 |
| 301.73 | 69.56 | 1.86 |
| 299.95 | 70.42 | 1.86 |
| 1426.84 | 81.66 | 4.39 |
| 193.61 | 66.87 | 1.65 |
| 433.91 | 69.81 | 2.14 |
| 351.93 | 72.21 | 1.91 |
| 249.10 | 69.21 | 1.70 |

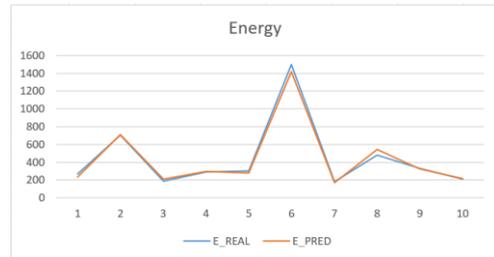

Fig. 5. Representation of real Energy vs predicted Energy values using Multiple Linear Regression Model.

- Mean Absolute Error: 30.07
- Accuracy: 90%

The predicted values of Energy was considered as satisfying since 90% of the values are inside the range of ± 15% of the real value (Fig.6). For a comparison purpose, we calculate also the mean-absolute-error between the predicted and real Energy values. The loss value will be compared to the rest of models that will be presented in the next sections.

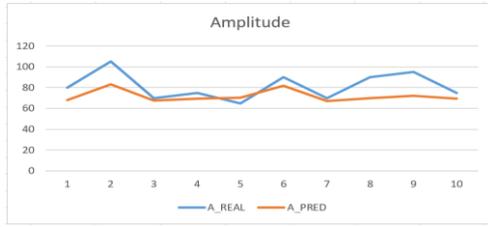

Fig. 6. Representation of real Amplitude vs predicted Amplitude values using Multiple Linear Regression Model.

- Mean Absolute Error: 10.73
- Accuracy: 70%

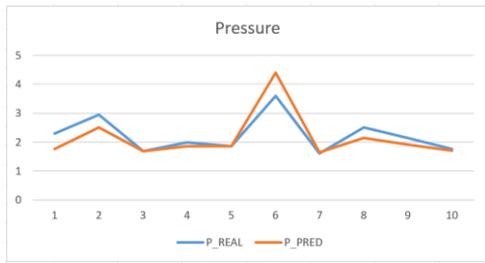

Fig. 7. Representation of real Pressure vs predicted Pressure values using Multiple Linear Regression Model.

- Mean Absolute Error: 0.26
- Accuracy: 50%

*B. Support Vector Regression*

The goal in this section is to apply the concept of Support Vector Machine for regression purpose, that means the response variable is not a categorical variable but a quantitative numerical variable. We are trying to do numerical prediction using a set of attributes and to find the relationship between the *n*-dimensional real vector attribute *X* and the *p*-dimensional response variable *Y* (while minimizing an error). This method consist of searching for the vectorial function *f(X)* which has at most, a deviation ε with respect to the training data, and which is flat as possible (complexity)[13].

To apply Multi-output SVR algorithm to our data, we started by defining and tuning of the standard hyper parameters which are: C=35, Kernel=Radial Basis Function, ε=0.1 and γ=0.025.

We performed then the prediction test for the same data as previous section (new unseen data), which gave us the result below (Tab.II).

TABLE II. PREDICTED PARAMETERS USING SUPPORT VECTOR REGRESSION MODEL

| Energy (Ws) | Amplitude (%) | Pressure (Bar) |
|---|---|---|
| 266.58 | 73.81 | 2.12 |
| 938.99 | 94.67 | 3.44 |
| 390.16 | 76.52 | 2.07 |
| 330.58 | 76.71 | 2.29 |
| 351.03 | 80.27 | 2.33 |
| 1038.22 | 83.57 | 3.50 |
| 234.03 | 68.62 | 1.87 |
| 358.35 | 76.98 | 2.26 |
| 463.77 | 80.78 | 2.30 |
| 783.68 | 90.72 | 2.63 |

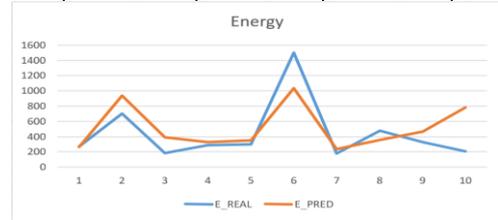

Fig. 8. Representation of real Energy vs predicted Energy values using SVR Model.

- Mean Absolute Error: 187.6
- Accuracy: 20%

The loss value is significant for Energy parameter compared to multiple linear regression model (which has MAE=30,07), also we can see that only 20% of prediction results are inside the range of [-15%,+15%] compared to real Energy values (Fig.8).

During the model tuning, different values of 'C' was tested, the previous prediction results concern the best C value for our data, whih is C=35.

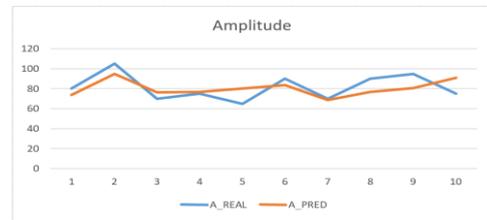

Fig. 9. Representation of real Amplitude vs predicted Amplitude values using SVR Model.

- Mean Absolute Error: 9.08
- Accuracy: 80%

For the Amplitude parameter prediction, our SVR model perform better than the Multiple Linear Regression model in term of loss value as well as the accuracy value since 80% of the predicted value are inside the defined range (Fig.9).

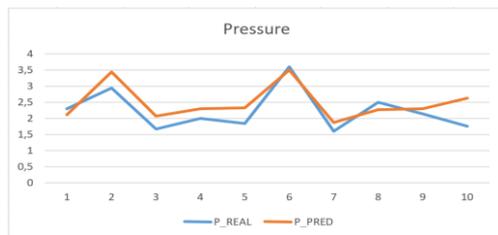

Fig. 10. Representation of real Pressure vs predicted Pressure values using SVR Model.

- Mean Absolute Error: 0.35
- Accuracy: 40%

## C. Artificial Neural Networks

Artificial neural networks was studied and described in multitude research work. In short description, the goal of ANN is to predict a Y-output (a characteristic) through a set of input Xi data, which are called observations. One of the ways to achieve this, highlighted by the research [14], was to simulate the response of an "artificial" neuron to these observations and to develop an algorithm to process and weight the observations to predict a characteristic. We will not develop the theoretical part of Artificial Neural Networks since it is deeply covered in other research works [15]. Our goal is find and apply this state of the art algorithms to new areas and achieve better development of the concerned field.

For our case study, we fed our data to different model architecture and checked the Mean Absolute Error value as well as the prediction result for a new input data.
Our selected model architecture is described as below (Tab.III):

TABLE III. ARCHITECTURE AND HYPERPARAMETERS OF THE ANN MODEL

| Number of inputs | 15 |
|---|---|
| Number of hidden layers | 1 |
| Number of outputs | 3 |
| Neurons in the hidden layer | 128 |
| Activation function | Rectifier Linear Unit |
| Learning rate | 0.003 |
| Regularization | 35% |
| Optimizer | Stochastic Gradient Descent |

TABLE IV. TREDICTED PARAMETERS USING ARTIFICIAL NEURAL NETWORKS MODEL

| Energy (Ws) | Amplitude (%) | Pressure (Bar) |
|---|---|---|
| 398.03 | 68.06 | 1.97 |
| 512.58 | 90.66 | 3.18 |
| 429.90 | 72.73 | 2.11 |
| 408.48 | 70.51 | 2.13 |
| 410.00 | 72.35 | 2.27 |
| 530.85 | 87.01 | 3.41 |
| 395.34 | 65.92 | 1.81 |
| 415.69 | 70.36 | 2.01 |
| 429.35 | 76.27 | 2.36 |
| 496.58 | 86.47 | 2.60 |

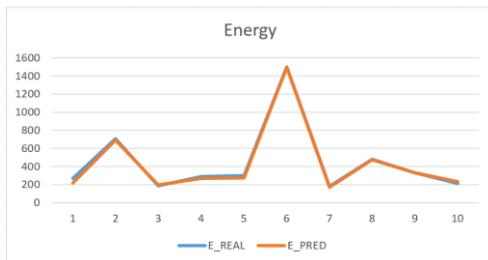

Fig. 11. Representation of real Energy vs predicted Energy values using Artificial Neural Networks model.

- Mean Absolute Error: 16.25
- Accuracy: 90%

Our Artificial Neural Networks model shows a good accuracy with the lowest loss value compared to both previous algorithms (Fig.11), which make it the best model in Energy prediction for new data, this is also explained by the capacity of generalization obtained from the weight regularization layer that we added before the output layer.

This model shows a stable behavior after 3000 iterations (approximatly 15 secondes of training).

The prediction of Amplitude values should be improved since it still lower than both previous algorithms (Fig.12).

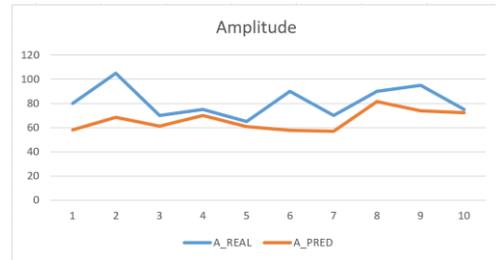

Fig. 12. Representation of real Amplitude vs predicted Amplitude values using Artificial Neural Networks model.

- Mean Absolute Error: 15.41
- Accuracy: 40%

For the Pressure parameter prediction (Fig.13), the ANN model shows also the lowest loss value for the new data compared to both previous methods. Even that, the multi-output regression model still performing the best accuracy for this prediction.

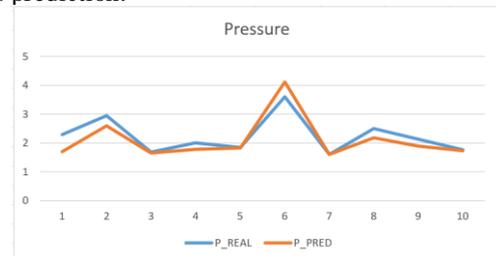

Fig. 13. Representation of real Pressure vs predicted Pressure values using Artificial Neural Networks model.

- Mean Absolute Error: 0.24
- Accuracy: 40%

## V. NEW APPROACH OF CONVOLUTIONAL NEURAL NETWORKS IMPLEMENTATION FOR PROCESS PARAMETERS PREDICTION

In this section, we will experiment the application of Convolutional Neural Networks algorithm that is mainly used for image recognition and classification [16][17], to a new field which is based on numerical data input and output. In our case, both input and output are initially numerical values (used in previous section with regression models).

The training input data represents the product characteristics (in our case wire cross sections) for an existing good quality finished product dataset, and the labels (output data) are the process parameters values (in our case Energy, Amplitude and Pressure) that are used to produce correctly this product and in respect to quality requirement.

The approach consist of converting the input data to a gray scaled pixels that will form a 2D image. Since our maximum input values are 15 (case of unilateral welding node), we decided to accept up to 16 input value for each product. For the product that are composed by less than 16 wires, we set a value of zero in the remaining columns (Fig.14).

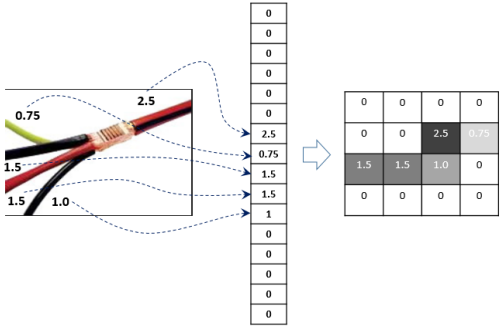

Fig. 14. Conversion of products characteristics into 2D matrix

For this specific problem, the position of wires is not considered due to their small impact on the result, which allow us to generate more 2D images from the same product by random permutation of matrix elements that we considered as data augmentation step (Fig.15).

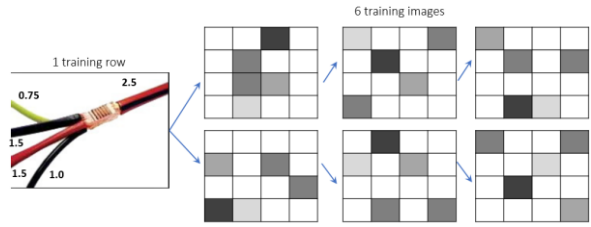

Fig. 15. Data augmentation by random permutation of matrix elements (6 new generations)

To allow different convolution operations (which lead to smaller image), we increased the scale of the input images from 4x4 pixels into 16x16 pixels using matrix interlaced replication .

After the data transformation to 2D matrix, data augmentation by elements permutation (6 times for each matrix), conversion of 2D matrix to gray scaled images and their size increase, our dataset was ready to feed our designed CNN model.

The model is composed from two convolution layers, one pooling layer and 2 fully connected layers. The detailed architecture is showed in Fig.16. The training outputs are kept as numerical values and we used a rectifier linear unit in the output layer in order to allow continuous output prediction. We included the batch normalization during the training phase with Stochastic Gradient Descent optimization.

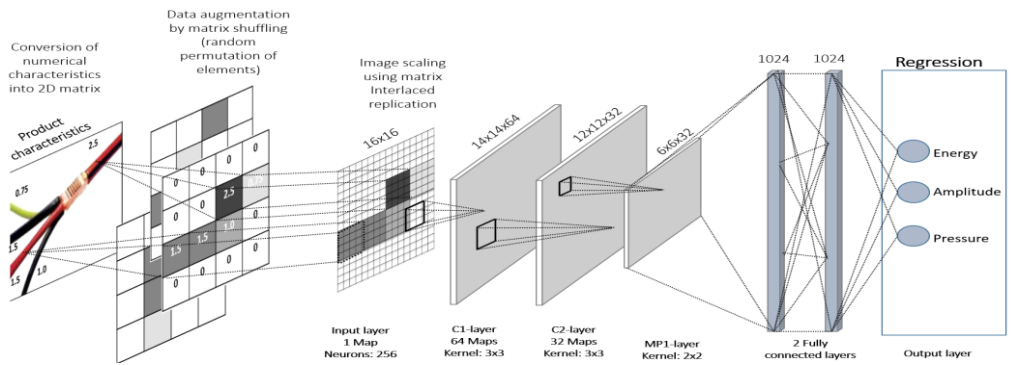

Fig. 16. The designed architecure of Convolutional Neural Networks model for industrial process parameters prediction based on product characteristics inputs.

A regularization layer (of type Dropout with p=35%) was used after the pooling layer in order to improve the generalization ability of our model and to avoid the over-fitting effect. The following figures (Fig.17, Fig.18, Fig.19) shows the prediction results for a new input data and the calculated error for each parameter. The CNN model is giving a good result on Energy prediction (close to the result obtained on the ANN model). The prediction accuracy for the Energy is 100% (inside +/-15% range) which is the best accuracy result for all the presented methods. Also, the result of Amplitude value prediction is more accurate than ANN model. We can notice that the model still need more parametrization in order to predict in a better way the Pressure and the Amplitude values (Tab.V).

TABLE V.    PREDICTED PARAMETERS USING CONVOLUTIONAL NEURAL NETWORKS MODEL

| Energy (Ws) | Amplitude (%) | Pressure (Bar) |
|---|---|---|
| 231.177 | 74.2211 | 2.00535 |
| 708.794 | 97.398 | 0.868664 |
| 210.754 | 71.2066 | 1.91596 |
| 294.413 | 72.5827 | 2.08099 |
| 279.429 | 76.2082 | 2.1896 |
| 1418.77 | 119.957 | 5.45809 |
| 168.135 | 62.4882 | 2.23339 |
| 543.234 | 73.8032 | 3.61656 |
| 324.864 | 69.0733 | 1.17486 |
| 218.227 | 72.7605 | 1.6103 |

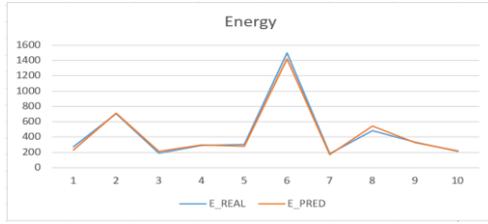

Fig. 17. Representation of real Energy vs predicted Energy values using CNN model.

- Mean Absolute Error: 26.2
- Accuracy: 100%

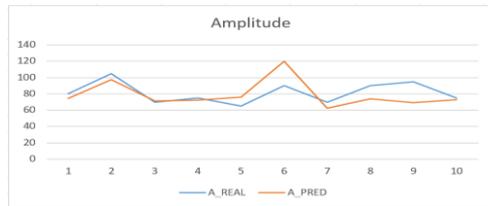

Fig. 18. Representation of real Amplitude vs predicted Amplitude values using CNN model.

- Mean Absolute Error: 11.0
- Accuracy: 60%

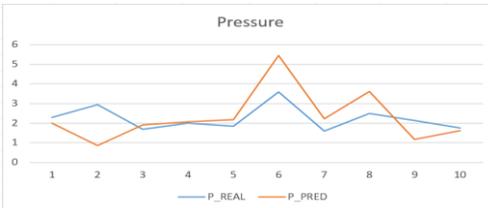

Fig. 19. Representation of real Pressure vs predicted Pressure values using CNN model.

- Mean Absolute Error: 0.78
- Accuracy: 30%

## VI. RESULT DISCUSSION

For a better evaluation of the models. a higher number of new input data should be concidered. In this paper. we considered only a set of 10 new products that we should produce on the ultrasonic welding machine. we performed the pridiction using our four models. the real output values are presented in the previous sections. In this section. we make a summary of the results obtained previously in term of loss (mean absolute error) and of accuracy. We remind that accuracy in our case mean that the predicted value is close by (+) or (-) 15% to the real value. In the real indusrial case, the accuracy of prediction is the best metric for model evaluation because it give us a direct idea about the faisability of the process parameters for a specific product.

### A. Energy prediction (Ws)

The best prediction accuracy for Energy parameter was obtained from the CNN model (Fig.20) that we designed according to our proposed approach in section 'V'. The mean absolute error for the same model is 26.2 Ws (Watt second), which is more significant than the ANN model. Since we care more about accuracy for our case study, we judge the CNN model as the best in Energy prediction.

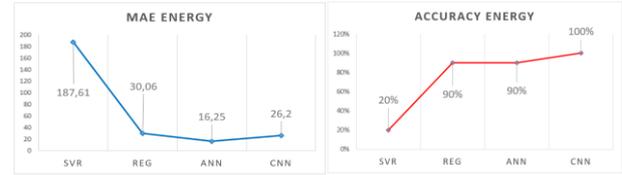

Fig. 20. Comparison of loss value and accuracy for energy prediction.

### B. Amplitude prediction (µm or %)

The Amplitude parameter was better predicted by the Support Vector Regression model (Fig.21). 80% of the predicted values are inside the tolerance range. The ultrasonic amplitude is usually measured in (µm), for our case it is represented in (%) which mean the mouvement position of the horn (0%= no mouvement; 100%= maximum horn amplitude). The model error is then 9.07%.

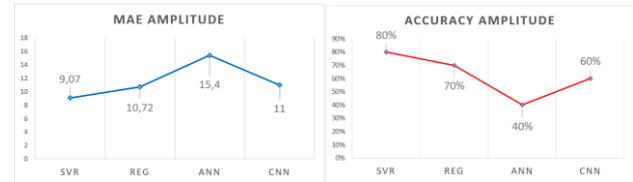

Fig. 21. Comparison of loss value and accuracy for amplitude prediction.

### C. Pressure prediction (bar)

The best accuracy of pressure prediction was obtained by the multi-regression model (Fig.22). this result still not satisfiying and should be improved since it is only 50%.

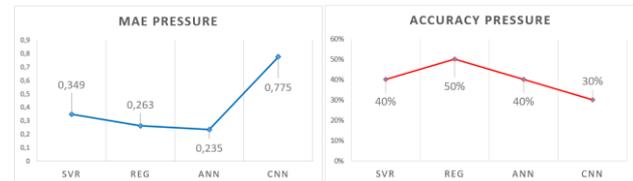

Fig. 22. Comparison of loss value and accuracy for pressure prediction.

### D. Welding tests using combination of models

In this part. we selected the best combination of predicted parameters based on previous results in order to make real tests on the welding machine. This helped us to validate the results. We made a new welding operations for the same set of products used in previous sections but this time using the predicted values as below:

- Energy: CNN model

- Amplitude: SVR model
- Pressure: Regression model

In below photos (Fig.23) we can see the tested products before and after welding.

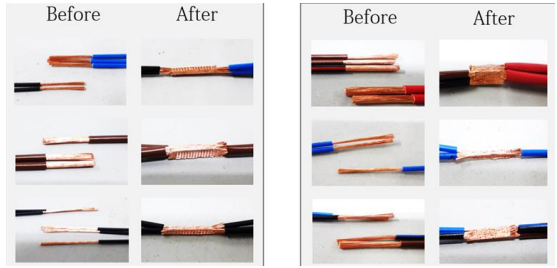

Fig. 23. Photos of testing samples after welding

The judgement of these tests was the same as the customer requirements. which are: the pull force value, the peel force value and the visual aspect of the welded node. For the ten welded product. 8 products was completely conform. 2 products was not acceptable as their peel force value is a bit under the limit (Tab.VI). also the visual aspect was not correct (damaged copper stand).

TABLE VI.  QAULITY EVALUATION OF THE WELDED SAMPLES

| Visual judgement | Pull force result (N) | Pull force treshhold (N) | Peel force result (N) | Peel force treshhold (N) |
|---|---|---|---|---|
| OK | 85 | 76 | 25 | 16 |
| OK | 500 | 311 | 118 | 87 |
| OK | 85 | 76 | 20 | 16 |
| OK | 213 | 201 | 43 | 45 |
| OK | 170 | 151 | 31 | 31 |
| OK | 465 | 311 | 90 | 87 |
| N.OK | 104 | 101 | 19 | 20 |
| OK | 115 | 101 | 23 | 20 |
| OK | 83 | 76 | 31 | 16 |
| OK | 106 | 101 | 26 | 20 |

VII. CONCLUSION

In this study, we presented two new approachs that can be further deployed in the industrial process field.

The first practice is to predict the process parameters for a new product taking their main characteristics as input (dimension, type, material etc.). For the training data. we used a list of different products that was previously produced in the same process and we defined their characteristics as training inputs. and their process parameters as training output. Then we built different machine learning algorithms to learn the relationship between the products characteristics and the process parameters. As demonstration. we applied this practice on ultrasonic welding process of copper wires. We concluded that for our case study. different models and algorithms can be used combinly to predict different parameters type.

The second contribution that we intended to introduce in this paper is an approach to use Convolutional Neural Networks to predict industrial process parameters following the same practice that we presented previoustly. We demonstarated the way to built a CNN model that can predict correctly the process parameters based on products characteristics. We also presented a generalization metholgy that can be applied to any similar problem.